\documentclass[12pt]{l4dc2020} 

\title[Identifying Mechanical Models of Unknown Objects with Differentiable Physics]{Identifying Mechanical Models through Differentiable Simulations}
\usepackage{times}

\author{%
 \Name{Changkyu Song} \Email{cs1080@rutgers.edu}\\
 \Name{Abdeslam Boularias} \Email{abdeslam.boularias@rutgers.edu}\\
 \addr Computer Science Department, Rutgers, the State University of New Jersey, Piscataway, NJ, USA.
}

\usepackage{url}
\usepackage{xcolor,colortbl}
\usepackage{amsmath,amsfonts,amssymb,wasysym}
\usepackage{verbatim}
\usepackage{mathrsfs}
\let\chapter\section
\usepackage{wrapfig}
\usepackage{graphicx} 
\usepackage{xspace}
\usepackage{enumitem}
\usepackage{capt-of}
\usepackage{algorithm2e}
\usepackage{multirow,multicol}
\usepackage{color}
\usepackage{soul}

\begin{document}

\maketitle
\begin{abstract}
 This paper proposes a new method for manipulating unknown objects through a sequence of non-prehensile actions that displace an object from its initial configuration to a given goal configuration on a flat surface. The proposed method leverages recent progress in differentiable physics models to identify unknown mechanical properties of manipulated objects, such as inertia matrix, friction coefficients and external forces acting on the object. To this end, a recently proposed differentiable physics engine for two-dimensional objects is adopted in this work and extended to deal forces in the three-dimensional space. The proposed model identification technique analytically computes the gradient of the distance between forecasted poses of objects and their actual observed poses, and utilizes that gradient to search for values of the mechanical properties that reduce the reality gap.
 Experiments with real objects using a real robot to gather data show that the proposed approach can identify mechanical properties of heterogeneous objects on the fly.
\end{abstract}

\section{Introduction}
 
 In robotics, nonprehensile manipulation  can be more advantageous than the traditional pick-and-place approach when an object cannot be easily grasped by the robot~\cite{PackingICRA2019}. For example, combined pushing and grasping actions have been shown to succeed where traditional grasp planners fail, and to work well under uncertainty by using the funneling effect of pushing~\cite{Dogar2011AFF}. Environmental contact and compliant manipulation were also leveraged in peg-in-hole planning tasks~\cite{GuanVR18}. Nonprehensile actions, such as pushing tabletop objects, have also been used for efficient rearrangement of clutter~\cite{Dogar2012,king2015,king2016,KingICRA2017}. 
 The problem of pushing a single target object to a desired goal region was previously solved through model-free reinforcement learning~\cite{Pinto-abs-1810-10654,DBLP:conf/icra/YuanSKWH18,BoulariasAAAI2015}, but this type of methods still requires colossal amounts of data and does not adapt rapidly to changes in the environment. We consider here only model-based approaches.
 
 The mechanics of planar pushing was explored in the past~\cite{Mason86}, and large datasets of images of planar objects pushed by a robot on a flat surface were also recently presented~\cite{fazeli2017ijrr,bauza2019iros}. While this problem can be solved to a certain extent by using generic end-to-end machine learning tools such as neural networks~\cite{bauza2019iros}, model identification methods that are explicitly derived from the equations of motion are generally more efficient. For instance, the recent work by~\cite{JJZhou2018} presented a simple and statistically efficient model identification procedure using a sum-of-squares convex relaxation, wherein friction loads of planar objects are identified from observed motions of objects. Other related works use physics engines as black-boxes, and global Bayesian optimization tools for inferring mechanical properties of objects, such as friction and mass~\cite{Shaojune2018IROS}. 
 
 \begin{wrapfigure}{r}{0.35\textwidth} 
    \includegraphics[width=\linewidth]{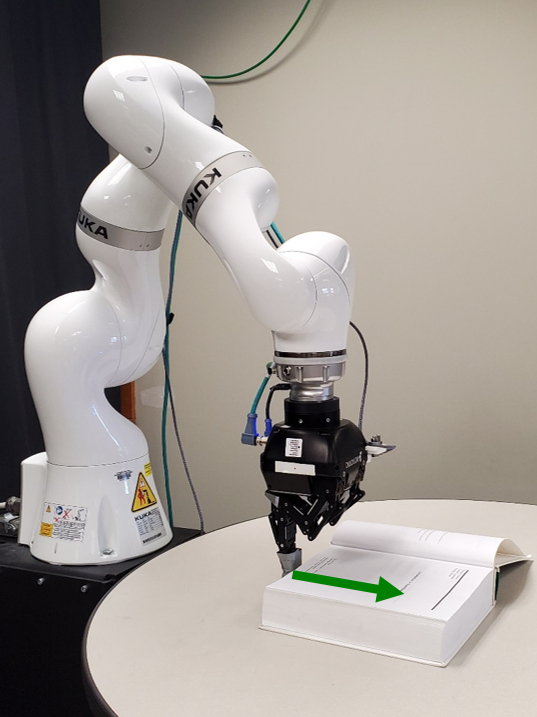}
    \caption{Robotic setup used in the experiments}
    \label{fig:robot}
    \vspace{-.68cm}
\end{wrapfigure}

 An alternative approach is to directly differentiate the simulation error with respect to the model of the object's parameters, and use standard gradient descent algorithms 
to search for optimal parameters. An advantage of this approach is the computational efficiency resulting from the guided search.
Unfortunately, most popular physics engines do not natively provide the derivatives along with the 6D poses of the simulated rigid bodies~\cite{ErezTT15,DART,PhysX,Bullet,ODE}. 
The only way to differentiate them is to use finite differences, which is expensive computationally. 

In the present work, we consider the general problem of sliding an unknown object from an initial configuration to a target one while tracking a predefined path. We adopt the recent differentiable physics engine proposed in~\cite{Belbute-Peres2017}, and extend it to take into account frictional forces between pushed objects and their support surface. To account for non-uniform surface properties and mass distributions, the object is modeled as a large number of small objects that may have different material properties and that are attached to each other with fixed rigid joints. A simulation error function is given as the distance between the centers of the objects in simulated trajectories and the true ones. The gradient of the simulation error is then used to steer the search for the mass and friction coefficients. 


\section{Related Work}
\emph{Classical system identification} builds a dynamics model by
minimizing the difference between the model's output signals and real-world response data for the same input signals~\cite{swevers1997optimal,Ljung:1999:SIT:293154}. Parametric rigid body dynamics models have also been combined with non-parametric model learning for approximating the inverse dynamics of a robot~\cite{nguyen2010using}. 
\emph{Model-based reinforcement learning} explicitly learns the unknown dynamics, often from scratch, and searches for an optimal policy accordingly~\cite{Dogar_2012_7076,LunchMason1996,isbell:physics:2014,ZhouPBM16,abbeel2006using}.
Particularly, {\it Gaussian Processes} have been widely used to model dynamical systems~\cite{Deisenroth:2011fu,Calandra2016,MarcoBHS0ST17,bansal2017goal,pautrat2017bayesian}. Black-box, derivative-free, Covariance Matrix Adaptation algorithm (\texttt{CMA})~\cite{hansen2006cma} is also shown to be efficient at learning dynamical models~\cite{chatzilygeroudis:hal-01576683}. {CMA} was also used for automatically designing open-loop reference
trajectories~\cite{tan2016simulation}, wherein trajectory optimization was performed in a physics simulator before real-world execution and real robot trajectories were used to fine-tune the simulator's parameters, the actual focus being the gains on the actuators. 
\emph{Bayesian optimization} (\texttt{BO})~\cite{shahriari2016taking} is a black-box approach
that builds a probabilistic model of an objective function that does not have a known closed form. A {GP} is often used to model the unknown function. 
{BO} has been frequently used in robotics for direct policy optimization ~\cite{pautrat2017bayesian}, such as gait optimization~\cite{Calandra2016}.
{BO} could also be used to optimize
a policy while balancing the trade-off between simulation and real-world experiments~\cite{MarcoBHS0ST17}, or to learn a locally
linear dynamics model~\cite{bansal2017goal}. BO was also successfully used to identify mechanical models of simple tabletop objects from a few images~\cite{Shaojune2018IJCAI}, and to search for robotic manipulation policies based on the identified models~\cite{Shaojune2018IROS}.
There has been a recent surge of interest in developing \emph{natively differentiable physics engines}. For example,~\cite{DegraveHDW16} used the Theano framework~\cite{DBLP:journals/corr/Al-RfouAAa16} to develop a physics engine that can be used to differentiate control parameters in robotics applications. The same framework  can be altered to differentiate model parameters instead.
A combination of a learned and a differentiable simulator was used to predict action effects on planar objects~\cite{DBLP:journals/corr/abs-1710-04102}. 
A differentiable framework for integrating fluid dynamics with deep networks was also used to learn fluid parameters from
data, perform liquid control tasks, and learn policies to manipulate liquids~\cite{Schenck2018SPNetsDF}.
Differentiable physics simulations were also used for manipulation planning and tool use~\cite{18-toussaint-RSS}. 
Recently, it has been observed that a standard physical simulation, which iterates solving the Linear Complementary Problem (LCP), is also differentiable and can be implemented in PyTorch~\cite{Belbute-Peres2017}. In~\cite{Mordatch:2012}, a differentiable contact model was used to allow for optimization of several  locomotion and manipulation tasks. 

\section{Problem Setup and Notation}

Parameters of material properties of objects, represented as a vector $\theta \in \Theta$ of dimension $D$, are given as input to the physics engine. For example, $\theta_{[0]}$ represents the object's density, while $\theta_{[1]}$ is the coefficient of friction between the object and a support surface. A real-world trajectory $\tau^*$ is a state-action sequence $(x_0,u_0, \dots, x_{T-1}, u_{T-1}, x_{T})$, where $x_t$ is the state of the pushed object, and $u_t$ is the action (force) applied at time $t$. 
In this work, we focus on the problem of pushing an unknown rigid object. The object is approximated as a finite set of elements. The object is thus composed of several connected cells $1,2,\dots,k$. Each cell $i$ has its own mechanical properties that can be different from other cells. State $x_t$ is a vector in $[SE(2)]^k$ corresponding to translation and rotation in the plane for each of the $k$ cells.
A corresponding simulated trajectory $\tau$ for assumed physical
parameters $\theta$ is obtained by starting at an initial state
$\hat{x}_0$, which is set to be the same as the initial state of the
corresponding real trajectory, i.e., $\hat{x}_0 = x_0$, and applying
the same control sequence $(u_0, u_1, \dots, u_{T-1})$. Thus,
the simulated trajectory $\tau$ results in a state-action sequence
$(\hat{x}_0, \hat{v}_0,u_0, \hat{x}_1,\hat{v}_1,u_1, \dots, \hat{x}_{T-1},\hat{v}_{T-1}, u_{T-1}, \hat{x}_{T})$,
where $\hat{x}_{t+1} = \hat{x}_{t} + \hat{v}_{t}dt$ is the next state
predicted by the physics engine using the model $\theta$. 
Predicted velocity $\hat{v}_{t}$ is a vector in $[SE(2)]^k$ corresponding to translation  and angular velocities in the plane for each of the $k$ cells. Predicted velocity $\hat{v}_{t+1}$ is given by 
$\hat{v}_{t+1} = V(\hat{x}_{t},\hat{v}_{t}, u_t, \theta), \hat{v}_{0} = \mathbf{0}$.
The goal is to identify model parameters $\theta^*$ that result in
simulated state outcomes $\hat{x}_{t+1}$ that are as close as possible to the real observed states $x_{t+1}$.  In other words, the objective is to solve the following optimization problem:

\begin{eqnarray}
\theta^* = \arg \min_{\theta \in \Theta} loss(\theta)  \stackrel{def}{=}  \sum_{t=0}^{T-2}\|x_{t+2} - \big(\hat{x}_{t+1} +  V(\hat{x}_{t},\hat{v}_{t}, u_t, \theta)dt \big) \|_2.
\label{simulationError}
\end{eqnarray}
In the following, we explain how velocity function $V$ is computed.

\section{Foundations}
We adopt here the formulation presented by~\cite{Belbute-Peres2017}, and we extend it to include frictional forces between a pushed object and a support surface. 
The transition function is given as $x_{t+1} = x_t + v_{t} dt$ where $dt$ is the duration of a constant short time-step. Velocity $v_{t}$ is a function of pushing force $\mu_t$ and mechanical parameters $\theta$. 
To find $v_{t+dt}$, we solve the system of equations of motion given in Figure~\ref{equation_motion}, where $x_t$ and $v_{t}$ are given as inputs, $[a,\rho,\xi]$ are slack variables, $[v_{t+dt},\lambda_e,\lambda_c,\lambda_f,\eta]$ are unknowns, and
$[\mathcal{M}, \mu_f] \stackrel{def}{=} \theta$ are the mechanical properties of the manipulated object  that are hypothesized. $\mathcal{M}$ is a diagonal mass matrix, where the diagonal is given as $[\mathcal{I}_1,\mathcal{M}_1,\mathcal{M}_1,\mathcal{I}_2,\mathcal{M}_2,\mathcal{M}_2,\dots, \mathcal{I}_k,\mathcal{M}_k,\mathcal{M}_k]$, where $\mathcal{I}_i$ is the moment of inertia of the $i^{th}$ cell of the object, and $\mathcal{M}_i$ is its mass. $\mu_F$ is a diagonal matrix containing the magnitudes of the frictional forces exerted on each cell of the pushed object in the direction opposite to its motion.

$\mathcal{J}_e (x_t)$ is a global Jacobian matrix listing all constraints related to the joints of the object. These constraints ensure that the different cells of the object move together with the same velocity, since the object is assumed to be rigid. $\mathcal{J}_c (x_t)$ is a matrix containing constraints related to contacts between the cells of the object. These constraints enforce that rigid cells of the object do not interpenetrate. $c$ is a vector that depends on the velocities at the contact points and on the combined restitution parameter of the cells. More detailed descriptions of $\mathcal{J}_e (x_t)$ and $\mathcal{J}_c (x_t)$ can be found in the supplementary material of~\cite{Belbute-Peres:2018:EDP:3327757.3327820}.

$\mathcal{J}_f$ is a Jacobian matrix related to the frictional forces, and it is particularly important for the objectives of this work. We will return to $\mathcal{J}_f$ in Section~\ref{friction}.
To present the solution more concisely, the following terms are introduced.
\begin{eqnarray*}
\alpha=-v_{t+dt}, \beta = \lambda_e, A = \mathcal{J}_e(x_t) , q = -\mathcal{M}v_t - dt \mu_t\\
\gamma =\begin{bmatrix} \lambda_c \\ \lambda_f \\ \eta \end{bmatrix}, 
s = \begin{bmatrix} a \\ \rho \\ \xi \end{bmatrix}, 
m = \begin{bmatrix} c \\ 0 \\ 0 \end{bmatrix}, 
s = \begin{bmatrix} a \\ \rho \\ \xi \end{bmatrix}, 
G = \begin{bmatrix} \mathcal{J}_e(x_t) & 0 \\ \mathcal{J}_f & 0 \\ 0 & 0 \end{bmatrix},
\mathcal{F} = \begin{bmatrix} 0 & 0 & 0 \\ 0 & 0 & E \\ \mu_f & -E^T & 0 \end{bmatrix}
\end{eqnarray*}
The linear complementary problem (LCP) becomes
\begin{eqnarray}
\begin{bmatrix} 
0  \\
s \\ 
0 
\end{bmatrix}
+
\begin{bmatrix} 
\mathcal{M} & G^T & A^T \\
G & \mathcal{F} & 0 \\ 
A & 0 & 0
\end{bmatrix}
\begin{bmatrix} 
\alpha  \\
\beta \\ 
\gamma 
\end{bmatrix}
=
\begin{bmatrix} 
-q  \\
m \\ 
0 
\end{bmatrix}
\label{lcp}
\end{eqnarray}
subject to $s\geq \mathbf{0}, \gamma \geq \mathbf{0}, s^T\gamma = 0$. 
\begin{figure}
{\scriptsize
\begin{equation*}
\boxed{
\begin{bmatrix} 0 \\ 0 \\ a  \\ \rho \\ \xi \end{bmatrix}
- \begin{bmatrix} \mathcal{M} & -\mathcal{J}_e(x_t) & -\mathcal{J}_c(x_t)  & -\mathcal{J}_f & 0 \\ 
\mathcal{J}_e(x_t) & 0 & 0 & 0 & 0\\ 
\mathcal{J}_c(x_t) & 0 & 0& 0 & 0 \\ 
\mathcal{J}_f & 0 & 0 & 0 & E \\ 
0 & 0 & \mu_{f} & -E^T & 0\end{bmatrix}
\begin{bmatrix} v_{t+dt} \\ \lambda_e \\ \lambda_c  \\ \lambda_f \\ \eta \end{bmatrix}
=
\begin{bmatrix} \mathcal{M}v_{t}+dt \mu_t \\ 0 \\ c  \\ 0 \\ 0 \end{bmatrix}
\textrm{s.t.}
\begin{bmatrix} a \\ \rho \\ \xi  \end{bmatrix} \geq 0,
\begin{bmatrix} \lambda_c \\ \lambda_f \\ \eta  \end{bmatrix} \geq 0,
\begin{bmatrix} a \\ \rho \\ \psi  \end{bmatrix} \begin{bmatrix} \lambda_c \\ \lambda_f \\ \eta  \end{bmatrix} = 0
}
\end{equation*}
}
\vspace{-0.5cm}
\caption{Equations of Motion}
\label{equation_motion}
\end{figure}

The solution is obtained, after an initialization step, by iteratively minimizing the residuals from the equation above through variables $\alpha,\beta,\gamma$ and $s$. The solution is obtained by utilizing the convex optimizer of~\cite{MattingleySB12}. 



\vspace{-1cm}
\section{Friction Model}
\label{friction}
We describe the Jacobian matrix $\mathcal{J}_f$ related to the Coulomb frictional forces between the object's cells and the support surface, and the corresponding constraints. This model is based on the one derived in~\cite{cline}. 
\vspace{-0.15cm}
\begin{eqnarray*}
\small 
\mathcal{D} = 
\begin{bmatrix} 
0 & 1 & 0 \\
0 & -1 & 0 \\
0 & 0 & 1 \\
0 & 0 & -1 
\end{bmatrix} , 
\mathcal{J}_f = 
\begin{bmatrix} 
\mathcal{D} & 0 & \dots & 0\\
0 & \mathcal{D} & \dots & 0\\
\vdots & \vdots & \dots & 0 \\
0 & 0 & \dots & \mathcal{D}
\end{bmatrix} ,
\mu_f = 
\begin{bmatrix} 
\mu_{f_1} & 0 & \dots & 0\\
0 & \mu_{f_2} & \dots & 0\\
\vdots & \vdots & \dots & \vdots \\
0 & 0 & \dots & \mu_{f_k}
\end{bmatrix} ,
E = 
\begin{bmatrix} 
\zeta & 0 & \dots & 0\\
0 & \zeta & \dots & 0\\
\vdots & \vdots & \dots & \vdots \\
0 & 0 & \dots & \zeta
\end{bmatrix}
\end{eqnarray*}
\vspace{-0.15cm}
$\zeta = [1,1,1,1]^T$, $\mathcal{J}_f$ and $E$ are both a $k\times(4k)$ matrix where $k$ is the number of cells in the object. $\mu_{f_i}$ is the unknow friction coefficient of cell $i$ with the support surface. In the original model of~\cite{cline}, matrix $\mathcal D$ was defined as 
$
\mathcal{D} = 
\begin{bmatrix} 
(p_{contact} \times d) &  d \\
(p_{contact} \times -d) &  -d 
\end{bmatrix},
$
where $d$ is a vector pointing to the tangential to a contact, and $-d$ is a vector pointing to the opposite direction. $p_{contact}$ is the contact point on the object. In our model, we consider only pushing actions that slide objects forward without rolling them over. We thus eliminate the first column of $\mathcal{D}$ which are multiplied by the angular velocities and replace it with a zero column. 
The friction terms have complementary constraints that are related to the contact points. We will omit the derivations here (see~\cite{cline}), but interactions between the objects and the support surface give rise to the following constraints,
$\mu_f -E^T \lambda_f \geq 0, \mathcal{J}_f v_{t+dt} + \gamma E \geq 0, \rho^T \lambda_f = 0, \xi^T \eta = 0.$



\section{Proposed Algorithm}

To obtain parameters $\theta^* = [\mathcal{M}^*,\mu_{f}^*]$, a gradient descent on loss function in Equation~\ref{simulationError} is performed, wherein the gradients are computed analytically. A first approach to compute the gradient is to use the {\it Autograd} library for automatic derivation in Python. We propose here a second simpler and faster approach that exploits three mild assumptions: 1) the manipulated object is rigid, 2) the contact point between the robot's end-effector and the object remains constant during the pushing action, and 3) collision between the robot's end-effector and the object is perfectly inelastic with a zero restitution coefficient. That is, the end-effector remains in contact with the object during the pushing action. The relatively low velocity of the end-effector (around 0.2 m/s) ensures the inelastic nature of the collision. The equation of motion can be written as:
\vspace{-0.1cm}
\begin{eqnarray}
 \mathcal{M} v_{t+dt}  -\mathcal{J}_e(x_t) \lambda_{e} -\mathcal{J}_c(x_t) \lambda_{c}  - \mu_{f} \lambda_c= \mathcal{M}v_{t}+dt \mu_t.
\end{eqnarray}
Thus, $V(\hat{x}_{t},\hat{v}_{t}, u_t, \theta) = v_{t+dt}   = \mathcal{M}^{-1} \big( \mathcal{J}_e(x_t) \lambda_{e} + \mathcal{J}_c(x_t) \lambda_{c}  +\mu_{f} \lambda_c + \mathcal{M}v_{t}+dt \mu_t \big).
$
Inverse mass matrix $ \mathcal{M}^{-1}$ exists because $\mathcal{M}$ is a diagonal full-rank matrix. The gradients of predicted velocity with respect to $\mu_f$ is give as:
$
\nabla_{\mu_f} V(\hat{x}_{t},\hat{v}_{t}, u_t, \theta) = \mathcal{M}^{-1} \lambda_c,
$
because $\nabla_{\mu_f} \mathcal{J}_e(x_t) \lambda_{e} = \nabla_{\mu_f} \mathcal{J}_c(x_t) \lambda_{c} = 0$ from assumptions 1-3.
The gradient of the loss in Equation~\ref{simulationError} is given by 
$\nabla_{\mu_f} loss(\theta) = \sum_{t=0}^{T-2} \nabla_{\mu_f} \|x_{t+2} - \big(\hat{x}_{t+1} +  V(\hat{x}_{t},\hat{v}_{t}, u_t, \theta)dt \big) \|_2 
= \sum_{t=0}^{T-2} \Big( x_{t+2} - \big(\hat{x}_{t+1} +  V(\hat{x}_{t},\hat{v}_{t}, u_t, \theta)dt \big)\Big) \mathcal{M}^{-1} \lambda_c dt = C_f \sum_{t=0}^{T-2} \Big( x_{t+2} - \big(\hat{x}_{t+1} +  \underbrace{V(\hat{x}_{t},\hat{v}_{t}, u_t, \theta)}_{\textrm{forward simulation}}\Big)$
where $C_f$ is a constant diagonal matrix. The real value of $C_f$ is of little importance as it will be absorbed by the learning rate of the gradient descent algorithm. A similar derivation for $\mathcal M$ gives $\nabla_{\mathcal M} loss(\theta) = C_m \sum_{t=0}^{T-2} \Big( x_{t+2} - \big(\hat{x}_{t+1} +V(\hat{x}_{t},\hat{v}_{t}, u_t, \theta)\Big)$,
\RestyleAlgo{boxruled}
with $C_M = \alpha C_f^{-\frac{1}{2}}$ and $\alpha$ is a constant factor. Thus, we use update rates $\alpha_{\textrm{rate}}$ and $\sqrt{\alpha_{\textrm{rate}}}$ for frictional forces magnitudes and mass matrix respectively, as as shown in Algorithm~\ref{algo}.
  \begin{algorithm}[h]
\KwIn{Real-world trajectory data $\tau^* = \{(x_{t},\mu_{t}, x_{t+1})\}$ for $t=0, \dots,T-1$, wherein  $x_t$ is a vector in $[SE(2)]^k$ corresponding to translation and rotation in the plane for each of the $k$ cells of a pushed unknown object, and $\mu_{t}$ is a force described by the contact point between the end-effector and one of the object's cells in addition to the pushing direction.
Predefined learning rate $\alpha_{\textrm{rate}}$, and loss threshold $\epsilon$.}
\KwOut{Parameters vector $\theta  = [\mathcal{M}, \mu_f]$}
Initialize $\mu_{f}$ randomly;

\Repeat{$loss \leq \epsilon$}{
$loss \leftarrow0$;

\For{$t\in\{0,T-2\}$}{
   Simulate $\{(\hat{x}_{t+1},\mu_{t+1})\}$ by solving the LCP in Equation~\ref{lcp} with parameters $\theta$ (or by using an off-the shelf physics engine) and get the predicted next state $\hat{x}_{t+2} = \hat{x}_{t+1} + V(\hat{x}_{t},\hat{v}_{t}, u_t, \theta)$;
   
   $loss \leftarrow loss + \|\hat{x}_{t+2} - x_{t+2}\|_2$;
   
   $\mu_{f} \leftarrow \mu_{f} + \alpha_{\textrm{rate}} (\hat{x}_{t+2} - x_{t+2});$
   
  $\mathcal{M} \leftarrow \mathcal{M} + \sqrt{\alpha_{\textrm{rate}}} (\hat{x}_{t+2} - x_{t+2});$

}
}
\caption{Learning Mechanical Models with Differentiable Physics Simulations}
\label{algo}
\end{algorithm}
\vspace{-0.5cm}

\section{Evaluation}
We report here the results of our experiments for evaluating the proposed method.
\vspace{-0.2cm}
\subsection{Baselines}
The compared baselines are \emph{ random search}, \emph{finite differences gradient}, \emph{automatic differentiation with Autograd}, and \emph{weighted sampling}. The weighted sampling search generates random values uniformly in the first iteration, and then iteratively generates normally distributed random values around the best parameter obtained in the previous iteration.
The standard deviation of the random values is gradually reduced over time, to focus the search on the most promising region.


\begin{figure}[t]
    \centering
    \begin{tabular}{cc}
 		\includegraphics[width=0.49\textwidth]{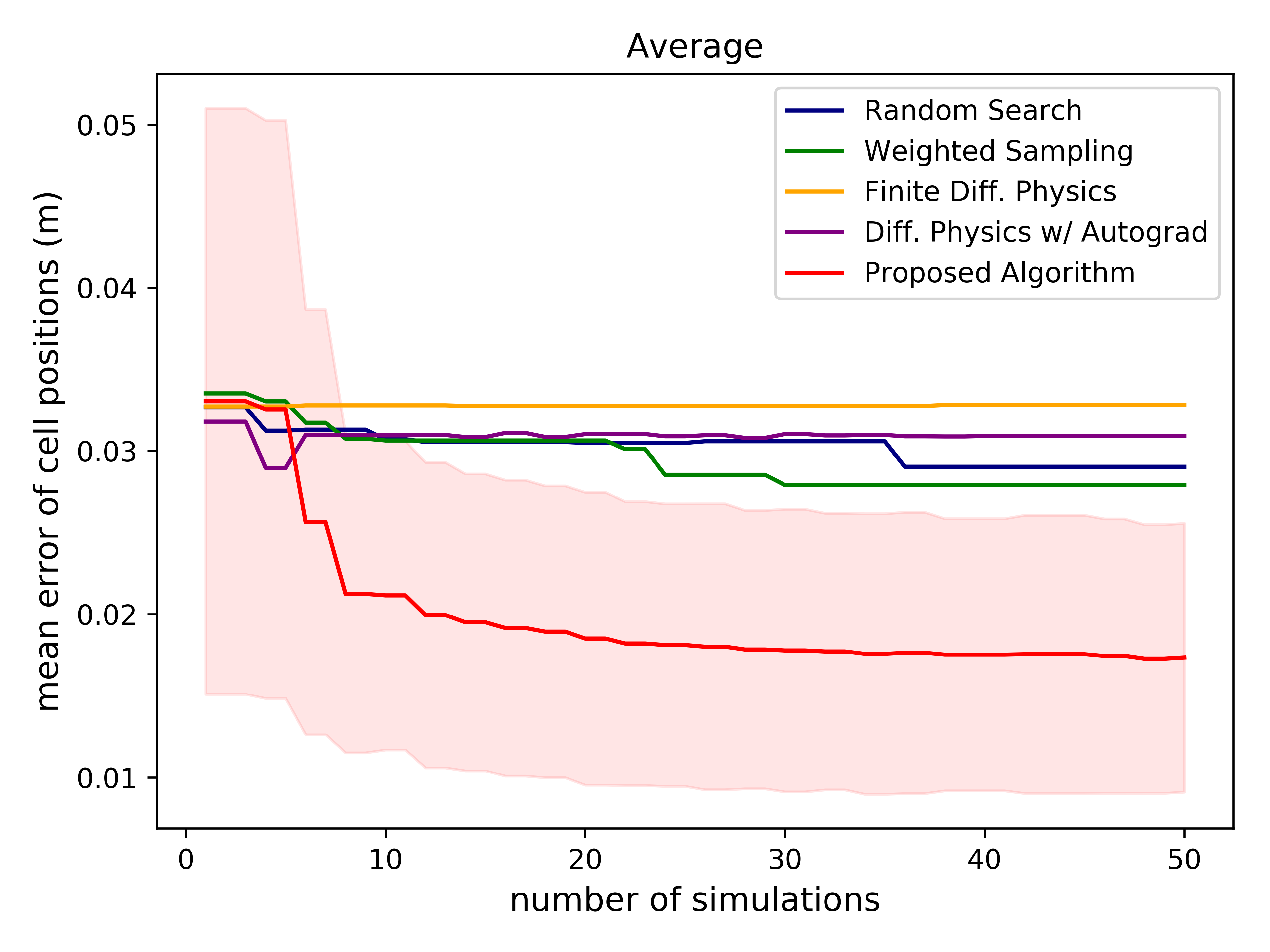} &
 		\includegraphics[width=0.49\textwidth]{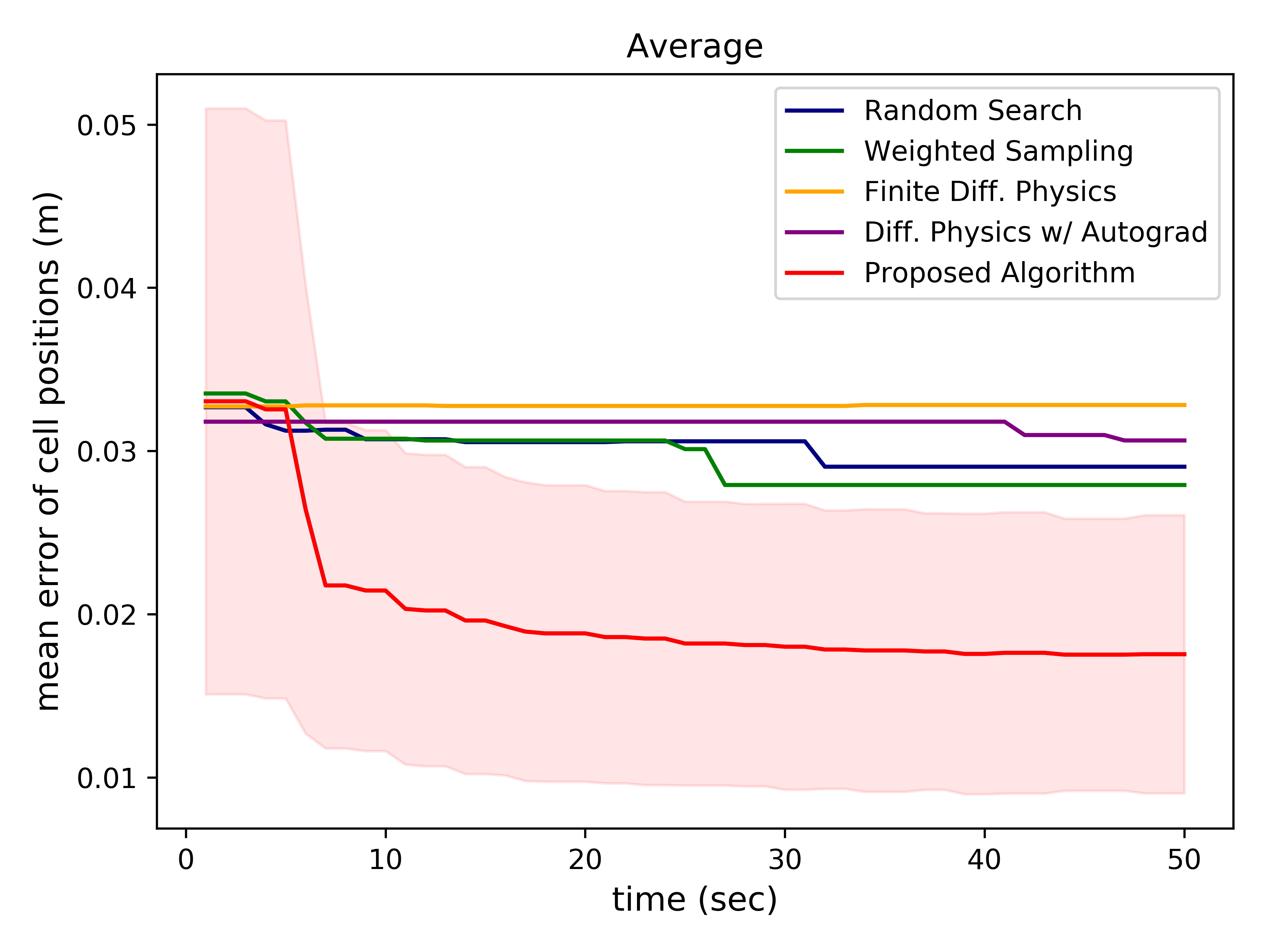}\\
    \end{tabular}
    \vspace{-0.5cm}
    \caption{Average predicted cell position error (in meters) over all objects and cells for each object as a function of the number of simulations (left) and the computation time (right).}
    \label{fig:quan_aver}
    \vspace{-0.8cm}
\end{figure}
\vspace{-0.2cm}
\subsection{Experimental Setup}
The experiments are performed on both simulated and real robot and objects. A rigid object is set on a table-top.
The robot's end effector is moved randomly to collide with the object and push it forward. The initial and final poses of the object are recorded.
The methods discussed above are used to estimate the object's mass and frictional forces that are distributed over its cells.
Since the ground-truth values of mass and friction are unknown, the identified models are evaluated in terms of the accuracy in the predicted pose of each cell, using a set of test data.
Five different objects are used in our experiments: a hammer, a ranch, a crimp, a toolbox, and a book.
A hammer has an unbalanced mass distribution because the iron head is much heavier than the wooden handle.
A crimp has high frictional forces on its heavy iron head and stiff handle.
A ranch is composed of the same material, however, its handle in the middle (main body) floats and does not touch the table, because of the elevated height of its side parts. Therefore, there are zero frictional forces on the handle. Finally, an open book that has a different number of pages on the left and right sides also has an unbalanced mass-friction distribution, resulting in rotations when pushed.
A toolbox also can have a various mass distribution depending on how the tools are arranged inside.

Note that our method does not assume that the full shape of the object is known, it uses only the observed top part and projects it down on the table to build a complete 3D geometric model. The geometric model is generally wrong because the object is seldom flat, but the parts that do not actually touch the table end up having nearly zero frictions in the identified model. Thus, identified low friction forces compensate for wrongly presumed surfaces in the occluded bottom part of the object, and the predicted motion of the object is accurate despite using inaccurate geometries.

In the simulation experiments, we simulated four random actions on each of the following objects for collecting training data: hammer, crimp, and ranch. We use the physics engine \textit{Bullet} for that purpose. The results are averaged over ten independent experiments, with a different ground-truth model of the object used in each experiment to generate data. The identified mass and friction models are evaluated by measuring the accuracy of the predicted motions on a test set of $12$ different random pushing actions.
In the real robot setup, we used a Kuka robotic arm and Robotiq 3-finger hand to apply the pushing actions, and recorded the initial and final object poses using a depth-sensing camera, as shown as Figure \ref{fig:robot}. The training data set contains only two random pushing actions, while the test data set contains five random pushing actions. The goal is to show how the robot can identify models of objects with a very small number of manipulation actions. The number of cells per object varies from $70$ to $100$ depending on the size of the object.

\begin{figure}[h]
    \centering
    \begin{tabular}{cc}
		\includegraphics[height=5.5cm]{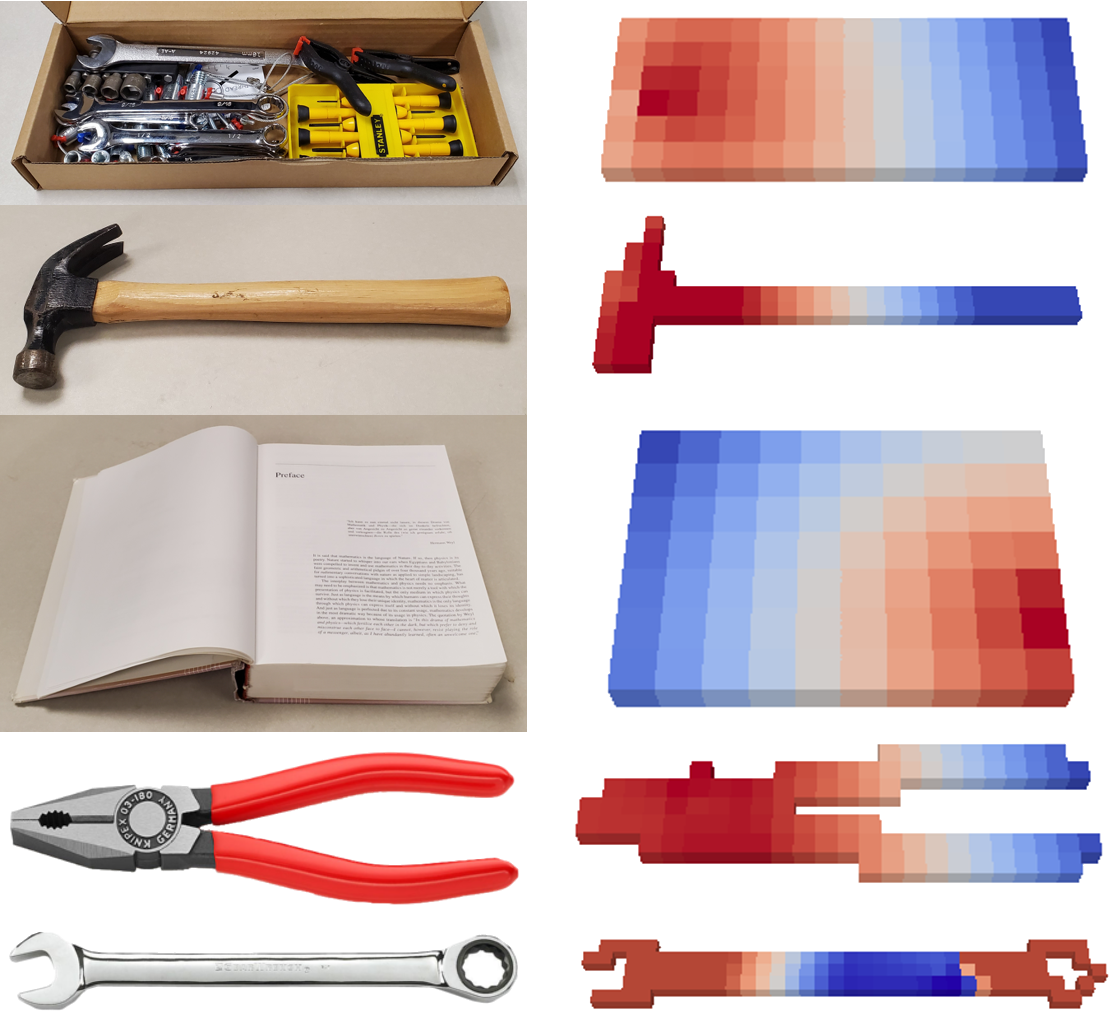}	         &
         \includegraphics[width=0.5\textwidth]{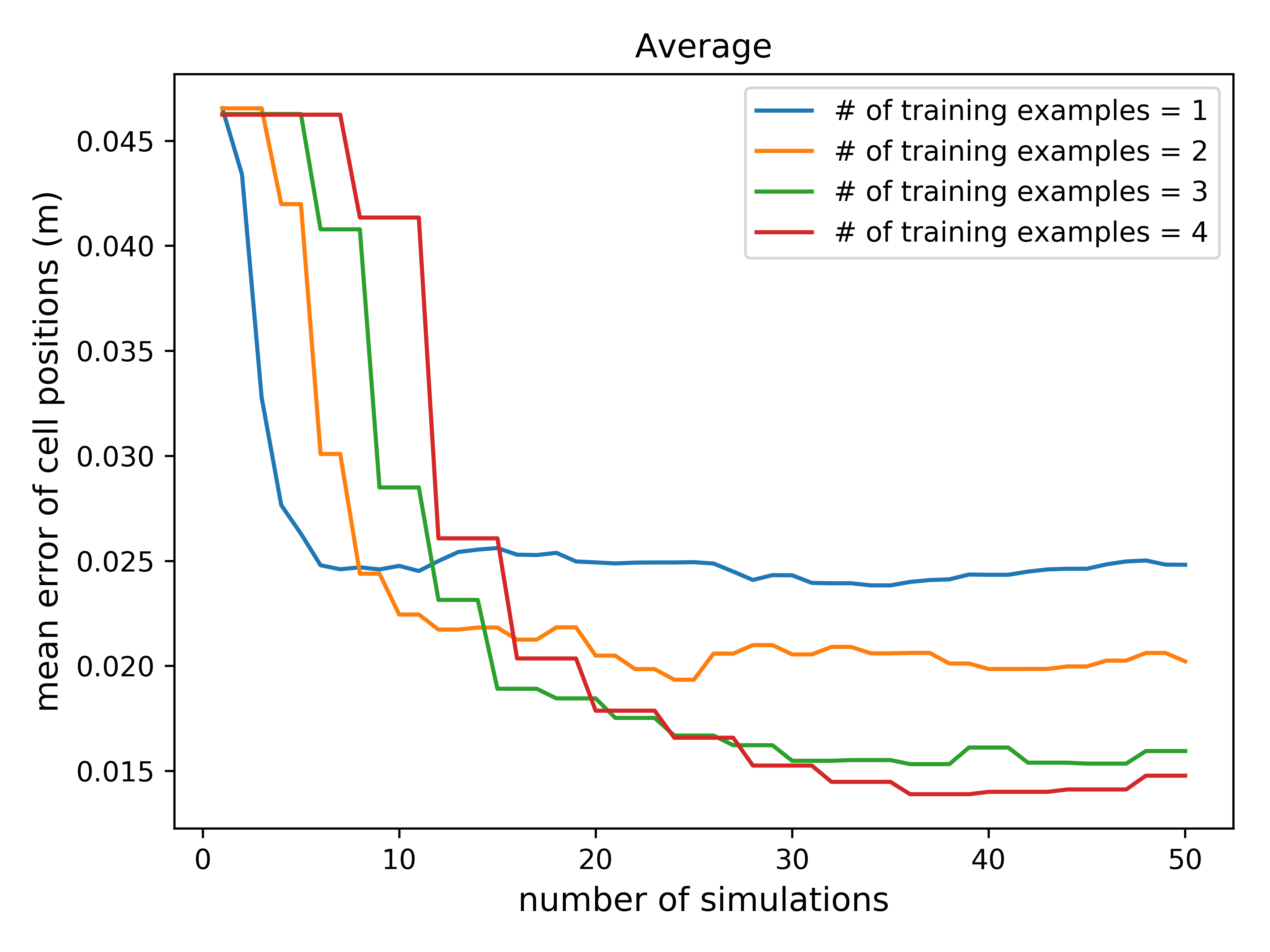}
		\\
		(a) & (b)
    \end{tabular}
    \vspace{-0.3cm}
    \caption{(a) Learned friction$\times$mass distributions. Red color means higher mass$\times$friction value while blue color means lower mass$\times$friction value. (b) Predicted cell position error with different numbers of training examples and simulations (gradient-descent steps).}
    \vspace{-0.8cm}
    \label{fig:quan_ntrain}
\end{figure}

\vspace{-0.2cm}
\subsection{Results}
Figure \ref{fig:quan_ntrain} (a) shows identified mass-friction distributions, given as the product of the mass and the friction coefficients of each cell. The mass-friction distributions of the first three objects are estimated in the real robot setup, and the last two are estimated in the simulation setup.
The toolbox contains heavy tools on the left side (ranches, bolts and nuts), while relatively light tools like plastic screw drivers and cramps are placed on the right side.
The proposed method was able to predict the unbalanced mass distribution of the box while it was covered. Likewise, the heavier iron head of the hammer and its light wooden handle are successfully estimated as well as the thicker and thinner sides of the book.
The proposed method successfully estimated the heavier part of the crimp, and the floating part of the ranch was simulated by much lighter friction values in the middle.

Figure \ref{fig:quan_aver} shows the difference between the predicted cell positions and the ground truth as a function of the number simulations used in the parameter estimations. The results first demonstrate that global optimization methods (random and weighted sampling) suffer from the curse of dimensionality due to the combinatorial explosion in the number of possible parameters for all cells. 
The results also demonstrate that the proposed method can estimate the parameters within a small number of simulations and a short computation time. The proposed algorithm was able to estimate the parameters with under $1.5cm$ average cell position error within $30$ seconds.
Surprisingly, the differential physics engine with \textit{Autograd} requires a significantly longer computation time as the number of cells increases, which is a critical in practice. The finite differences approach also failed to converge to an accurate model due to the high computational cost of the gradient computation, as well as the sensitivity of the computed gradients to the choice the grid size. 

Finally, Figure \ref{fig:quan_ntrain} (b) shows how the number of training actions improves the accuracy of the learned model. Increasing the number of training actions allows the robot to uncover properties of different parts of the object more accurately.
Using a larger number of training actions slows down the convergence of the gradient-descent algorithm, but improves the accuracy of the learned model.
\vspace{-0.7cm}
\vspace{-0.25cm}
\section{Conclusion}
\vspace{-0.1cm}
To identify friction and mass distributions of unknown objects pushed by a robot, we proposed a new method that consists in dividing an object into a large number of connected cells, with each cell having different mechanical properties. We adopted a differentiable physics engine that was recently proposed to simulate contact interactions between 2-dimensional objects, and we extended it to deal with frictional forces occurring on table-top 3D objects. In addition to the automatic derivation of the engine, based on {\it Autograd}, we presented a simple gradient-descent algorithm that exploits weak assumptions about the object and the collision to simplify the form of the gradient of reality-gap loss function with respect to the object's parameters. The proposed algorithm was tested in simulation and with real objects, and shown to be efficient in identifying models of objects with non-uniform mass-friction distributions. 

\noindent\textbf{Acknowledgments}{~This work was supported by NSF awards 1734492, 1723869 and 1846043.}

\bibliographystyle{abbrv}
\bibliography{bibliography}
\end{document}